\documentclass[letterpaper,10pt,conference]{ieeeconf}
\IEEEoverridecommandlockouts
\usepackage{newtxtext,newtxmath}
\usepackage{amsmath}
\usepackage{booktabs}
\usepackage{graphicx}
\usepackage{tabularx}
\usepackage{array}
\usepackage{url}
\usepackage{cite}
\usepackage[hidelinks]{hyperref}
\providecommand{\tightlist}{\setlength{\itemsep}{0pt}\setlength{\parskip}{0pt}}
\makeatletter
\def\bstctlcite#1{\@bsphack
  \@for\@citeb:=#1\do{%
    \edef\@citeb{\expandafter\@firstofone\@citeb}%
    \if@filesw\immediate\write\@auxout{\string\citation{\@citeb}}\fi}%
  \@esphack}
\let\sharedkvbibliography\thebibliography
\renewcommand{\thebibliography}[1]{%
  \sharedkvbibliography{#1}%
  \setlength{\itemsep}{-0.18em}%
  \setlength{\parsep}{0pt}%
}
\makeatother

\title{SharedKV-BT: Node-Local Typed Decisions for Behavior-Tree Agents}
\author{Naoki Wake$^{1}$ and Justin Wagle$^{1}$%
\thanks{$^{1}$Naoki Wake and Justin Wagle are with Microsoft.
{\tt\small naokiwake@microsoft.com, Justin.Wagle@microsoft.com}}}

\begin{document}
\maketitle

\begin{abstract}
Agent tasks require sequences of interdependent decisions. Autoregressive models support more flexible decision interfaces than conventional classifiers but incur the latency of token-by-token generation. Recent shared-prefix methods reduce this cost by reusing encoded context and scoring multiple decisions in parallel, but do not model decision dependencies or verify execution. We propose SharedKV-BT, where each active node of a behavior tree (BT) exposes stage-local fields and candidates, and Shared-KV scores the candidates in parallel and passes the selected decision to a separate execution system. We tested SharedKV-BT on robot manipulation, mobile navigation, and computer-use tasks. Across three tasks, SharedKV-BT made typed decisions \(2.36\)--\(4.15\times\) faster than prompt-matched autoregressive decoding. On the manipulation task, node-local Shared-KV improved joint decision accuracy from 75\% to 94\% and closed-loop success from 0\% to 60\%. Fixed-score policy replay showed that stage gating prevented out-of-order actions and external postconditions prevented premature completion.
\end{abstract}


\bstctlcite{IEEEtran:BSTcontrol}

\section{Introduction}\label{introduction}

Agents must make sequences of interdependent decisions to achieve a goal. A robot cannot place an object before grasping it, and a computer-use agent cannot paste a file before selecting its source and opening the destination. An appropriate action and its arguments, such as the target object or destination, need to be selected based on the current execution stage and environment state.

Conventional classifiers select from a predefined set of classes \cite{bishop2006pattern}. Open-set recognition extends this setting by rejecting inputs from classes not seen during training \cite{scheirer2013toward,geng2021recent}, but it still does not provide application-defined, stage-specific decision interfaces.
Autoregressive models support flexible interfaces specified through text \cite{brown2020language}, but sequential token decoding incurs inference latency that increases with output length~\cite{stern2018blockwise,leviathan2023fast}. More recently, typed-decision systems such as Jev ask a model to fill named fields by choosing among application-defined candidates, avoiding token-by-token generation \cite{typesafe2026system}.

Typed-decision interfaces, however, focus only on choosing the value of each field and do not manage execution itself \cite{typesafe2026system}. They do not define when a decision should be made, which candidates are valid at that point, or whether the chosen action succeeded. As a result, they can act out of order, repeat a completed action, or stop because the model claimed success. Behavior trees (BTs), on the other hand, are commonly used to structure agent execution through conditional branching and failure recovery \cite{iovino2022survey,colledanchise2017hybrid,colledanchise2014robustness}.

\begin{figure*}[t]
\centering
\includegraphics[width=\textwidth]{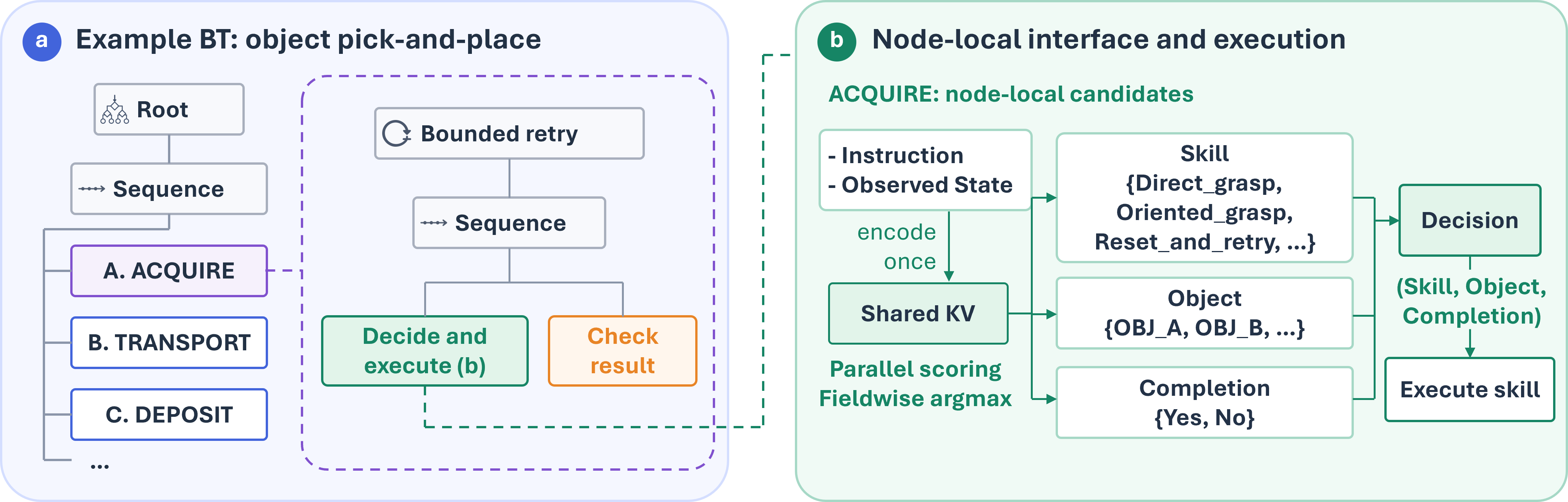}
\caption{SharedKV-BT schematic. (a) An example pick-and-place BT sequences \textsc{Acquire}, \textsc{Transport}, and \textsc{Deposit}; the active \textsc{Acquire} branch expands into a bounded retry loop over typed decision and execution followed by an external postcondition check. (b) The instruction and observed state are encoded once, Shared-KV scores the illustrative node-local skill, object, and completion candidates in parallel, and fieldwise argmax forms the decision. Completion is diagnostic; \textsc{Check Result} determines whether the active stage succeeded and controls BT progression.}
\label{fig:architecture}
\end{figure*}

In this paper, we propose SharedKV-BT, which combines a Jev-style typed-decision interface with BT control (Fig.~\ref{fig:architecture}). Each BT node defines the current execution stage and exposes only the fields and candidates that are valid at that stage. For example, in the schematic pick-and-place example, \textsc{Acquire}, \textsc{Transport}, and \textsc{Deposit} nodes expose different skills and arguments. This node-local design focuses the model on the decisions relevant to the current stage. The model encodes the instruction and observed state once as a task-state prefix and reuses the resulting key-value (KV) cache to score complete candidates in parallel. A separate execution system executes the selected action, while the BT's external postcondition checks the system state after execution and controls progression and retry. Section~\ref{tasks} specifies the experiment-specific interfaces.

Our findings answer three questions:

\noindent\textbf{Efficiency.} \emph{How much does shared-prefix reuse reduce decision cost?} On model inputs saved at decision points, Shared-KV was \(2.36\)--\(4.15\times\) faster than prompt-matched constrained autoregressive decoding; a controlled comparison using identical candidate batches isolated a \(1.29\)--\(2.09\times\) benefit from cache reuse alone (Section~\ref{decision-efficiency}).

\smallskip
\noindent\textbf{Focus.} \emph{Does exposing only the current node's choices improve decisions and execution?} Yes. Relative to an interface that exposed all fields and candidates at every decision (the global interface), node-local Shared-KV improved correct joint typed decisions from 75\% (90/120) to 94\% (113/120) and closed-loop completion from 0\% to 60\%  (Section~\ref{node-local-interface-routing}).

\smallskip
\noindent\textbf{Authority.} \emph{Which errors do stage gating and external postconditions prevent?} Stage gating prevented out-of-order actions, while external postconditions prevented model-reported completion from advancing an unsatisfied task (Section~\ref{control-of-candidate-sets-and-completion}).

The contributions of this paper are:

\begin{enumerate}
\def\labelenumi{\arabic{enumi}.}
\tightlist
\item
  An architecture in which each active BT node provides a typed-decision interface with predefined candidates, while a shared task-state prefix supports parallel candidate scoring.
\item
  Controlled evidence for the three corresponding mechanisms: prefix reuse for efficiency, node-local interface routing for focus, and stage gating with external postconditions for execution authority.
\item
  End-to-end evaluations across robot manipulation, mobile navigation, and computer-use tasks, with decision errors separated from downstream execution failures.
\end{enumerate}

\section{Related Work}\label{related-work}

\subsection{Robot Execution Architectures}\label{robot-execution-architectures}

Robot architectures separate fast action-level skills from higher-level decision making. Subsumption uses priority-ordered layers \cite{brooks1986robust}, whereas Gat's three-layer architecture \cite{gat1998three} and 3T \cite{bonasso1997experiences} place an executive between planning and real-time control. BTs formalize periodic updates (ticks) and success or failure returns \cite{marzinotto2014unified} and have been studied for sequential composition \cite{colledanchise2017hybrid}, robustness and safety \cite{colledanchise2014robustness}, and robotic applications \cite{iovino2022survey}. BT learning includes branch and tree design \cite{dey2013qlbt} and continuous action-node parameters \cite{iannotta2025adaptation}.

In a separate line of work, task and motion planning (TAMP) methods jointly search over discrete action sequences and continuous robot motions, propagating geometric failures back to task-level search~\cite{kaelbling2011hierarchical,garrett2021integrated}. SharedKV-BT instead operates at a different layer, selecting among candidates exposed by a predefined BT rather than planning action sequences or trajectories.

\subsection{Language-Model Planning and Skill Selection}\label{language-model-planning-and-skill-selection}

Large language models (LLMs) and vision-language models (VLMs) have generated or revised task procedures from prompts, demonstrations, fine-tuning, or heuristic planning \cite{cao2023robot,izzo2024btgenbot,cai2024hbtp,ao2025llmbtplanner}, including runtime-generated visual-condition nodes \cite{wake2025vlmdriven}, language-model action sequences \cite{wake2023chatgpt}, and multimodal plans from demonstrations \cite{wake2024gpt4vision}.

Reusable skill libraries separate task descriptions from hardware execution \cite{takamatsu2026skillagents}, and Plan-and-Act studies when to invoke an LLM \cite{sasabuchi2025planandact}. Other systems span free-form code or plan generation \cite{liang2023code,vemprala2024chatgpt,driess2023palme,huang2022zeroshot} and predefined skill selection: SayCan combines language-model scores with scores of whether an action is feasible \cite{ahn2022saycan}, ProgPrompt specifies available actions in program-like prompt context \cite{singh2023progprompt}, and Inner Monologue conditions later reasoning on feedback \cite{huang2022innermonologue}. Unlike these methods, SharedKV-BT combines fast candidate scoring through Shared-KV with stage-constrained decision making through the BT.

\subsection{Complete-Candidate Scoring and Shared-Prefix Inference}\label{complete-candidate-scoring-and-shared-prefix-inference}

Complete-candidate likelihood scoring is not novel \cite{brown2020language}; prior work has also examined alternative label wordings and prompt-dependent calibration \cite{holtzman2021surface,zhao2021calibrate}. Related structured-generation methods constrain outputs through token masking or schemas, including XGrammar, Outlines, and Structured Outputs \cite{dong2024xgrammar,willard2023guided,openai2024structured}, while Toolformer learns when to insert tool calls during generation \cite{schick2023toolformer}. Jev frames structured inference as typed decisions over application-defined candidates \cite{typesafe2026system}, and concurrent work scores candidate identifiers in parallel \cite{li2026llm2jev}.

Shared-prefix inference has also been studied at the systems level. PagedAttention manages KV caches \cite{kwon2023pagedattention}, SGLang reuses prefixes through RadixAttention \cite{zheng2024sglang}, Hydragen separates shared-prefix from candidate-specific computation \cite{juravsky2024hydragen}, and ArborKV manages caches for branching search \cite{chen2026arborkv}. SharedKV-BT applies shared-prefix reuse to multi-step agent execution, with each BT node exposing multiple typed fields and candidates that share a task-state prefix.

\section{SharedKV-BT System}\label{materials-and-methods}

SharedKV-BT can be applied to autoregressive models that provide a probability for each token and allow reuse of their KV caches. All primary experiments used the open-weight Qwen2.5-3B-Instruct \cite{qwen2024qwen25}. As an additional model-robustness check, we repeated Adaptive PickPlace with Qwen3-4B in non-thinking mode using its native chat template \cite{yang2025qwen3}.

\subsection{Shared-KV Complete-Candidate Scoring}\label{shared-kv-complete-candidate-scoring}

A typed decision fills named fields by choosing from predefined candidates; for example, a grasp node may expose \texttt{skill} = \{\texttt{GRASP}, \texttt{ORIENTED\_GRASP}\} and \texttt{operand} (action argument) = \{\texttt{CUBE\_A}, \texttt{CUBE\_B}\}. The model represents the common instructions and system state as text and encodes them as a task-state prefix \(x\). For each field \(f\), its name and candidates form prompt \(p_f\). If candidate \(c_i\) has tokens \((t_{i,1},\ldots,t_{i,m_i})\), its score within field \(f\) is

\[
s(c_i\mid f)=\frac{1}{m_i}\sum_{j=1}^{m_i}
\log P(t_{i,j}\mid x,p_f,t_{i,<j}).
\]

The KV cache for \(x\) is computed once and reused while the candidate completions for all fields are evaluated in one batch. Only candidate tokens contribute to the mean log score, which reduces preference for shorter labels and distinguishes labels sharing their first token. SharedKV-BT records mean and per-token scores and selects the highest-scoring candidate in each field.
These scores rank the supplied candidates; we do not treat them as reliable probabilities of correctness or as evidence that the candidate set contains a valid action.

Notably, independent fieldwise argmax does not, in general, guarantee cross-field compatibility. Arbitrary interfaces therefore require an explicit compatibility check. The node-local interfaces used in our experiments, however, are designed so that all exposed field combinations are executable.

To isolate the benefit of KV reuse, we compare four methods for answering the same typed decision. Autoregressive (AR) generation produces the answer one token at a time without constraining it to the candidate set. Prompt-matched constrained AR starts from the same task-state prefix, field suffix, and candidates as complete-candidate scoring, then selects the highest-probability next token at each step (i.e., greedy decoding) while a token trie (i.e., a prefix tree) blocks every sequence outside the candidate set. Sequential scoring evaluates each candidate but recomputes the task-state prefix for every candidate. Shared-KV encodes the task state once and reuses its KV cache to score all candidates in one batch.

\subsection{Behavior-Tree Decision Interface}\label{behavior-tree-decision-interface}

The BT determines when a decision is requested and exposes only the fields and candidates of the active node. The model selects their values but neither generates the procedure nor certifies the outcome. An external execution system executes the selected decision; its implementation is outside the scope of this paper. At each step, the system converts the state to text, scores the active interface, and executes the selected skill. An external evaluator then checks the postcondition, which the BT uses to determine the next execution step (Fig.~\ref{fig:architecture}).

In this implementation, each node exposes fields for a skill and its arguments. For the control experiments, each node additionally exposes a model-reported completion field. Under full BT control, neither model-reported completion nor the skill's return value authorizes progression. A separate evaluator advances the tree only after confirming the external postcondition from the underlying system state. The fixed-score policy replay holds all model scores constant while changing only the execution policy. In the condition without the external postcondition gate, model-reported completion instead controls progression (Section~\ref{experimental-conditions}).

\subsection{Tasks}\label{tasks}
The tasks serve complementary roles in evaluating SharedKV-BT: cross-domain tasks test end-to-end applicability and decision efficiency, while controlled task variants isolate node-local interface routing and execution control.
The primary execution comparisons cover five tasks across three environments: Robosuite Stack for manipulation \cite{zhu2020robosuite}, RoboCasa NavigateKitchen for navigation \cite{nasiriany2024robocasa}, and the Clock, Notepad, and File Explorer tasks in WindowsAgentArena \cite{bonatti2024windows}. Stack requires a dependent grasp-then-place sequence, while NavigateKitchen requires ordered navigation to five fixtures with recovery from an injected failure. A broader manipulation evaluation adds full PickPlace, NutAssemblySquare, and Door, yielding eight tasks overall, with 30 episodes for each manipulation task. Full PickPlace requires placing Milk, Bread, Cereal, and Coke Can in their corresponding bins.

Adaptive PickPlace extends the same full-PickPlace environment with three BT nodes repeated for each object: \textsc{Acquire}, \textsc{Transport}, and \textsc{Deposit}. The \textsc{Acquire} node selects among \texttt{DIRECT\_GRASP}, \texttt{ORIENTED\_GRASP}, and \texttt{RESET\_AND\_RETRY} from structured inputs describing object orientation and whether a previous grasp failed. \textsc{Transport} and \textsc{Deposit} expose \texttt{MOVE\_TO\_BIN} and \texttt{ALIGN\_AND\_PLACE}, respectively. Simulator postconditions, which are Boolean conditions computed from simulator state, verify grasp, transport, placement, and task completion.

\begin{figure}[t]
\centering
\includegraphics[width=\columnwidth]{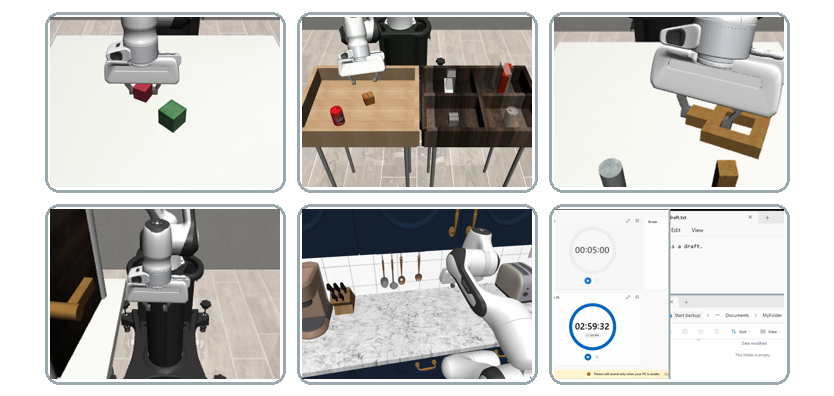}
\caption{Evaluated tasks. The top row shows Stack, full PickPlace, and NutAssemblySquare; the bottom row shows Door, NavigateKitchen, and a composite of Clock, Notepad, and File Explorer from WindowsAgentArena runs that passed the platform evaluator. Images illustrate the task interfaces rather than additional evaluation trials.}
\label{fig:overview}
\end{figure}

\subsection{Experimental Conditions}\label{experimental-conditions}

To separate the effects of inference method and execution control, we evaluate AR and Shared-KV under two execution policies on Stack and NavigateKitchen. Flat execution exposes actions from all stages without BT-based progression control, whereas BT execution exposes only node-local candidates and advances after its external postcondition is satisfied. This design yields four conditions, namely AR-flat, SharedKV-flat, AR-BT, and SharedKV-BT. For WindowsAgentArena, we run the full system four times per task, starting each run from a fresh application state, and check the result with the platform's own evaluator. For the broader manipulation suite (Stack, full PickPlace, NutAssemblySquare, and Door), we run the complete system end to end and check whether the robot's actions succeed in the simulator.

All four Stack and NavigateKitchen conditions share the model (Qwen2.5-3B-Instruct), observations, actions, skills, initial states, and evaluator. Separately, a fixed-score policy replay isolates the roles of stage gating and the external postcondition gate. It reuses identical saved Stack scores under four policies: full BT control with both gates, no stage gating, no external postcondition gate, and flat execution with neither gate. Because the model scores are fixed, differences in execution errors can be attributed to the execution policy.

Adaptive PickPlace compares node-local Shared-KV with a global-interface Shared-KV condition that always exposes all five skills, four objects, four destinations, and all typed fields. A secondary prompt-matched constrained AR condition receives exactly the node-local interface. The BT, decision points, physical skills, retry budget, seeds, interventions, and external postconditions are otherwise identical. Stage-inappropriate, wrong-object, and wrong-destination selections are rejected and recorded without correction.

\subsection{Evaluation Protocol}\label{evaluation-protocol}

\noindent\textbf{Latency evaluation.} To isolate inference latency, the latency experiments replay exact model inputs saved at decision points in Stack, NavigateKitchen, and Clock. Thus every method receives the same instruction, system state, fields, and candidates. Each Qwen2.5-3B-Instruct measurement uses two warm-up trials followed by ten measured trials. Measurements use one NVIDIA RTX 2080 Ti GPU with 11 GB of memory, FP16 arithmetic without quantization, scaled dot-product attention, PyTorch 2.7.1+cu126, Transformers 4.57.6, and CUDA 12.6. CUDA is synchronized around each measured procedure. The Stack, NavigateKitchen, and Clock inputs contain 139, 83, and 51 prefix tokens and 8, 11, and 11 field-candidate pairs, respectively.

The controlled cache-reuse comparison uses identical ordered candidate batches. The full-prefix batch repeats the prefix for every candidate, whereas Shared-KV computes the prefix once and reuses the expanded cache for the same candidate token sequences. Tokenization is excluded from both timers; tensor construction, transfer, model forward passes, scoring, and cache copying are included. A separate end-to-end benchmark starts before prompt construction and tokenization and compares Shared-KV, sequential scoring, prompt-matched constrained AR, and unconstrained AR. End-to-end timing includes Python-level token-trie masking, and only the matched-batch comparison isolates the effect of KV reuse. Five-repeat sweeps vary prefix length, field-candidate count, and candidate length.

\noindent\textbf{Execution evaluation.}
The four-condition Stack and NavigateKitchen evaluations use simulation seeds 0--9, and the recovery intervention attempts the same NavigateKitchen seeds after injecting the same planner rejection at the first goal. Pair validation retains all ten seeds with the same ordered goals across conditions. Only the recovery-enabled BT may select the \texttt{MAP\_REPLAN} recovery action. Each WindowsAgentArena task uses four runs from a fresh application state, assessed by the platform evaluator. The manipulation study runs 30 episodes each for Stack, full PickPlace, NutAssemblySquare, and Door using external simulator postconditions.

Adaptive PickPlace uses simulation seeds 0--29. For each seed, the four objects follow a rotated schedule with two direct-grasp states, one oriented-grasp state, and one reset-required state, yielding 60, 30, and 30 required states. The paired analysis replays the same 120 saved pre-execution states under all three interfaces. Because typed states repeat across seeds, the resulting counts are descriptive rather than 120 independent samples. A separate set of 30 closed-loop episodes per condition measures task success, completed object--bin relations, retries, rejected selections, and latency.

The candidate-order sensitivity analysis replays the saved Qwen2.5 \textsc{Acquire} states and the 90 global-interface \textsc{Transport} decision states under the published order, its reverse, and five fixed random permutations for each field. Each permutation changes only the candidate order; it does not change interface membership, state, prompt semantics, or scoring. Each fixed ordering is reported separately; predictions are neither averaged nor selectively reported based on favorable orderings.

We report latency, task success, completed stages, invalid or out-of-order actions, premature completion, and recovery. We evaluate decision validity separately from downstream execution. When methods produce the same valid decision, subsequent failures are attributed to execution rather than the decoder.

\section{Results}\label{results}

\subsection{Decision Efficiency}\label{decision-efficiency}

Holding candidate batching fixed, prefix reuse reduced decision-only latency from 207.1 to 98.9 ms on Stack, 208.6 to 119.5 ms on NavigateKitchen, and 170.4 to 131.9 ms on Clock. The resulting cache-reuse-only speedups were 2.09, 1.75, and 1.29 times, respectively (Table~\ref{tab:latency}).

\begin{table*}[t]
\centering
\caption{Qwen2.5-3B latency on model inputs saved at decision points. Matched-batch values are mean \(\pm\) 95\% confidence intervals over ten runs. End-to-end values are mean latencies from input preparation through decision output in the prompt-matched constrained AR comparison.}
\resizebox{\textwidth}{!}{%
\begin{tabular}{@{}lrrrrrr@{}}
\toprule
\textbf{Context} & \multicolumn{3}{c}{\textbf{Matched batch}} & \multicolumn{3}{c}{\textbf{End-to-end comparison}} \\
\cmidrule(lr){2-4}\cmidrule(l){5-7}
& \textbf{Full prefix (ms)} & \textbf{Shared-KV (ms)} & \textbf{Speedup} & \textbf{Shared-KV (ms)} & \textbf{Constrained AR (ms)} & \textbf{Speedup} \\
\midrule
Stack & \(207.1 \pm 0.6\) & \(98.9 \pm 0.5\) & \(2.09\times\) & 99.8 & 413.8 & \(4.15\times\) \\
NavigateKitchen & \(208.6 \pm 0.6\) & \(119.5 \pm 0.4\) & \(1.75\times\) & 121.4 & 286.0 & \(2.36\times\) \\
Clock & \(170.4 \pm 0.9\) & \(131.9 \pm 10.0\) & \(1.29\times\) & 134.2 & 318.8 & \(2.38\times\) \\
\bottomrule
\end{tabular}
}
\label{tab:latency}
\end{table*}

In the end-to-end implementation comparison, Shared-KV was 2.36--4.15 times faster than prompt-matched constrained AR. These end-to-end speedup factors include preprocessing, decoding, and Python-level token-trie masking and therefore are not estimates of cache reuse alone. Peak memory remained comparable across the tested implementations.

All constrained outputs were valid candidates. Constrained AR matched the Shared-KV selection on Stack but selected different valid candidates on NavigateKitchen and Clock. This disagreement reflects the objective: greedy tokenwise decoding need not select the candidate with the highest mean complete-sequence log likelihood. The comparison therefore uses the same predefined choices but does not assume that both methods select the same one.

\begin{figure}[t]
\centering
\includegraphics[width=\columnwidth]{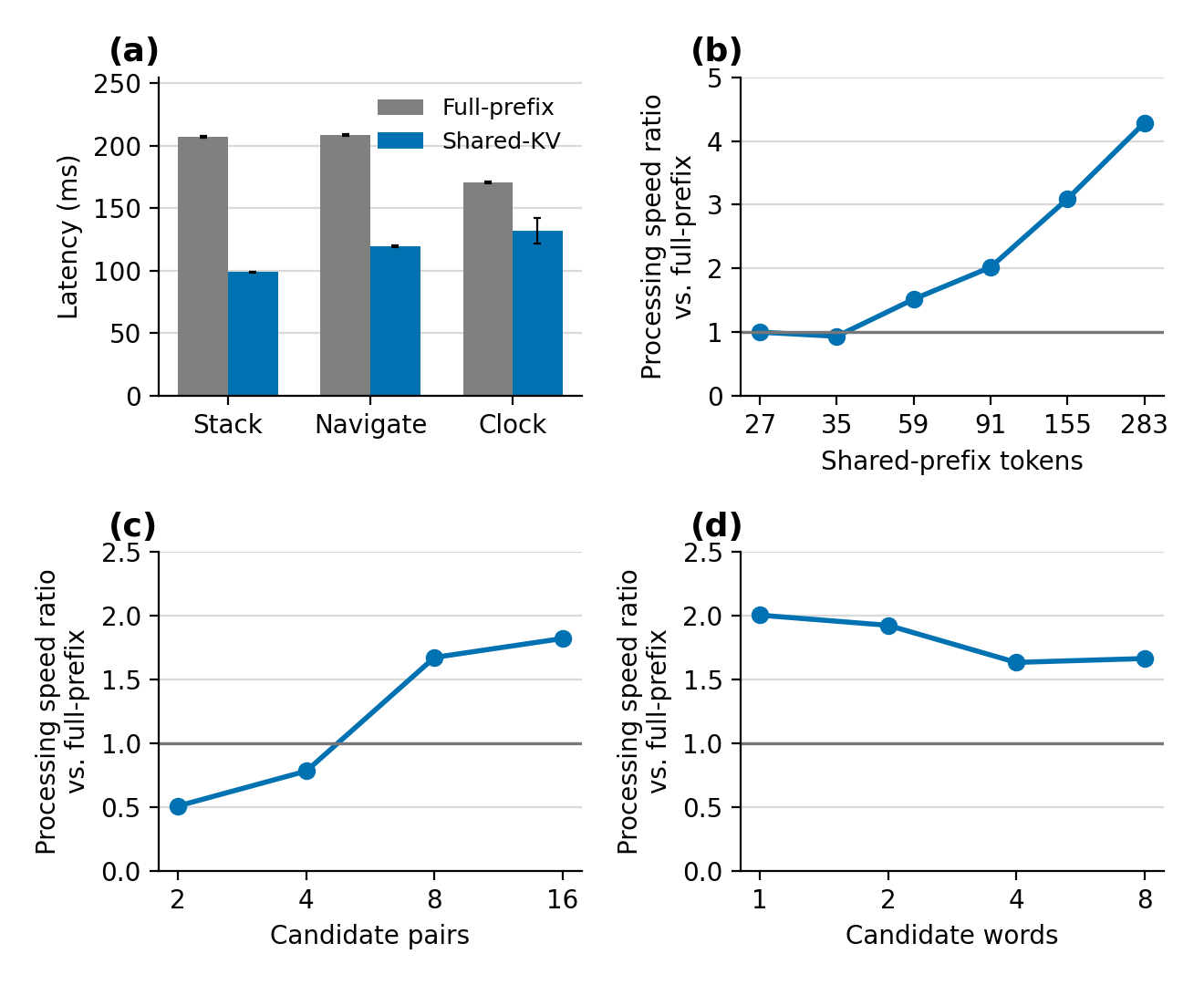}
\caption{Qwen2.5-3B matched-batch cache-reuse comparison. (a) Latency is compared using identical candidate batches with and without prefix reuse; error bars show 95\% confidence intervals over ten runs. The sweeps vary (b) prefix length, (c) the number of field-candidate pairs, and (d) candidate length. The dashed line indicates equal latency. Shared-KV is slower with only two candidates because cache-copy overhead outweighs the computation saved by prefix reuse, but becomes faster as more prefix computation is reused.}
\label{fig:kvreuse}
\end{figure}

Shared-KV and full-prefix scoring selected identical candidates in all Qwen2.5-3B latency trials, and their GPU memory use was comparable. The controlled sweeps identify the conditions under which reuse helps rather than claiming a universal speedup (Fig.~\ref{fig:kvreuse}). Speedup increases with prefix length, reaching 4.29 times at 283 tokens, and with the number of field-candidate pairs, falling to 0.51 when only two are scored. Across one- to eight-word candidates, speedup ranged from 1.63 to 2.01 times.

\subsection{Node-Local Interface Routing}\label{node-local-interface-routing}

\begin{figure}[t]
\centering
\includegraphics[width=\columnwidth]{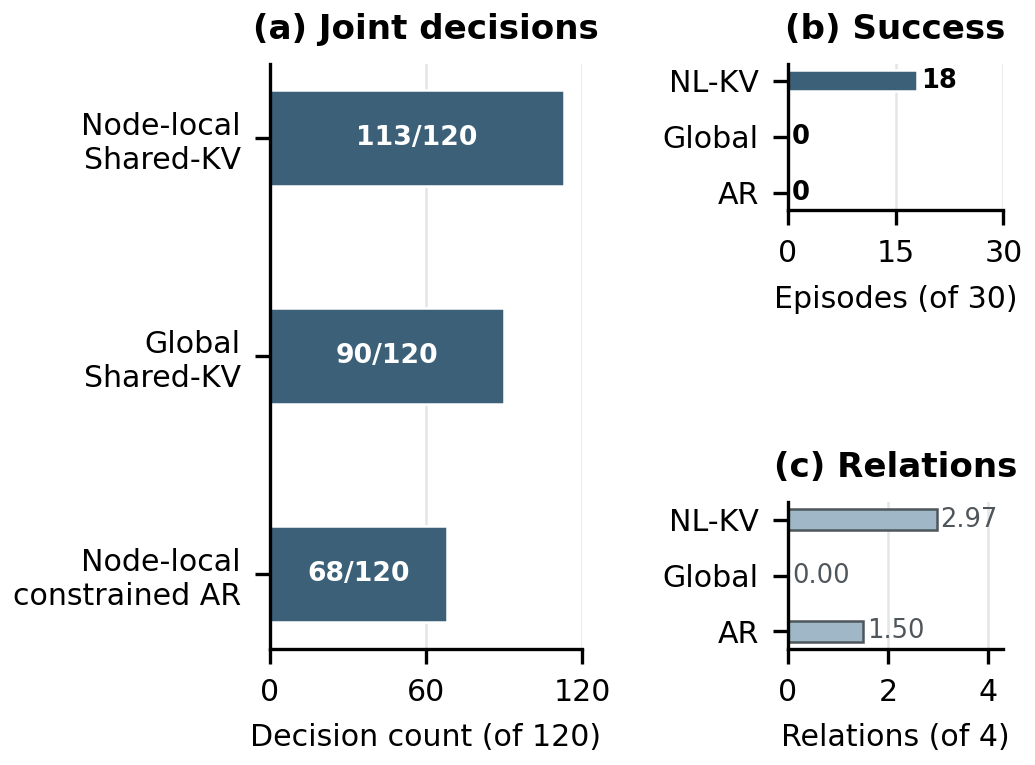}
\caption{Adaptive PickPlace isolates node-local interface routing. (a) Correct joint typed decisions on the predefined schedule of saved pre-execution states; counts are descriptive because the same state descriptions repeat across seeds. Closed-loop outcomes report (b) successful four-object episodes and (c) mean completed object--bin relations over 30 episodes per condition. NL-KV denotes node-local Shared-KV, Global denotes global-interface Shared-KV, and AR denotes node-local constrained AR. Global-interface selections were rejected rather than corrected.}
\label{fig:adaptive}
\end{figure}

Adaptive PickPlace directly tested whether the BT changes what the model must consider, not only when it decides. On the paired saved pre-execution states, node-local Shared-KV produced correct joint typed skill--object decisions in 113/120 states, compared with 90/120 for global-interface Shared-KV and 68/120 for node-local constrained AR (Fig.~\ref{fig:adaptive}). The global-interface decoder selected the required acquisition skill in every state but selected the wrong object in 30/120 states, whereas node-local Shared-KV made seven reset-state skill errors. Node-local interface routing therefore improved the complete typed decision, although it did not improve every field independently.

Node-local Shared-KV averaged 128 ms per paired decision, compared with 316 ms for the larger global interface and 625 ms for constrained AR. In closed loop, node-local Shared-KV completed 18/30 four-object episodes and averaged 2.97/4 object--bin relations. The global-interface condition completed 0/30 because, after a valid acquisition, it repeatedly selected an acquisition skill at the \textsc{Transport} node; all 90 such stage-inappropriate selections were rejected. Constrained AR also completed 0/30 and averaged 1.50/4 relations because it selected \texttt{DIRECT\_GRASP} in every oriented-grasp state. Thus the BT's node-local interface affected both decision quality and downstream execution.

The candidate-order sensitivity analysis showed that the node-local interface remained more accurate than the global interface across all tested orders, although individual predictions still changed with candidate order. Across the seven fixed orders, node-local Shared-KV produced 90--113/120 correct Acquire decisions, compared with 14--90/120 for the global interface, and node-local accuracy was higher under every order. At the saved \textsc{Transport} pre-execution states, node-local Shared-KV selected \texttt{MOVE\_TO\_BIN} in 90/90 states under every order, whereas the global interface produced a stage-inappropriate skill in 90/90 states under every order. Candidate order still changed the selected Acquire values in 46/120 node-local states and all 120 global states. The analysis therefore supports the node-local interface routing result across the tested orders while identifying order sensitivity in likelihood-based candidate scoring.

The additional Qwen3-4B check reproduced the same interface effect. Node-local and global-interface Shared-KV produced 120/120 and 90/120 correct joint typed decisions, respectively, with mean latencies of 189 and 480 ms. In a limited 10-seed closed-loop check, the node-local condition completed 8/10 episodes and averaged 3.5/4 relations, whereas the global-interface condition completed 0/10 and produced 30 rejected stage-inappropriate selections. These supplementary results do not replace the predefined Qwen2.5 evaluation set.

\subsection{Execution Authority under Fixed Scores}\label{control-of-candidate-sets-and-completion}

Stage gating and external postcondition verification prevented different classes of execution error. To isolate these mechanisms from model variation, we replayed the same scores at 15 saved grasp, placement, and completion states from Stack seeds 0--4 and changed only which actions and completion signals the policy allowed. Full BT control produced no errors. Without stage gating, the global action catalog, which contains actions from all stages, selected \texttt{DONE} before grasping and before placement in 10/15 states, making these selections out of order. Without the external postcondition gate, the model's completion output terminated the task at those same 10/15 states before the stack was complete. Flat execution exhibited both errors in the same 10/15 states. Because every condition used identical scores, the differences arose from the execution policy, not the model. This fixed-score policy replay isolates the effects of the gates at saved states; it does not estimate how often the same score patterns occur in new episodes.

\subsection{Task Success and Execution Errors}\label{end-to-end-execution}

We next test whether these mechanism-level effects persist in complete episodes.

On Stack, AR-BT and SharedKV-BT completed all 10 seeds without invalid, out-of-order, or premature-completion errors. Both flat conditions failed all 10 seeds and each produced 60 out-of-order actions. Within the BT conditions, Shared-KV reduced mean decision latency from 153.4 to 75.1 ms, a \(2.04\times\) speedup over AR.

On NavigateKitchen, both flat conditions failed all 10 episodes and produced 27 and 26 out-of-order actions for AR and Shared-KV, respectively. AR-BT and SharedKV-BT each completed 1/10 episodes with no invalid outputs, out-of-order actions, or premature completions and a mean of 0.5 of five goals completed. The two BT decoders selected identical slots for every seed. Shared-KV reduced mean decision latency from 688.8 to 334.9 ms, a \(2.06\times\) speedup. For each BT decoder, the nine unsuccessful episodes failed during map-route planning at the first goal, identifying navigation planning rather than decoder disagreement as the bottleneck. Results for all four conditions are reported in the supplementary evidence.

SharedKV-BT also completed all 12 WindowsAgentArena runs, with four runs from a fresh application state for each of Clock, Notepad, and File Explorer. The platform evaluators verified the running timer, exact saved text, and final folder location. This result demonstrates portability across the tested interfaces, not general computer-use reliability.

\subsection{Recovery Under Controlled Failure}\label{recovery-under-controlled-failure}

Beyond preventing invalid progression, we test whether the recovery branch restores progress after a controlled failure.

The NavigateKitchen intervention retained all ten paired runs with matching ordered goals. All ten recovery-disabled episodes stopped at the first goal, whereas Shared-KV selected the \texttt{MAP\_REPLAN} recovery action in all ten enabled episodes. Recovery restored first-goal progress in one episode, which completed all five ordered goals. The branch therefore activated consistently under the injected planner rejection, but downstream navigation still limited physical recovery and full-task completion.

\subsection{Boundaries of Constrained Selection}\label{boundaries-of-constrained-selection}

Complete-candidate scoring resolved labels that first-token scoring could not distinguish. On WindowsAgentArena-derived candidate sets containing observed first-token collisions, complete scoring matched 27/28 explicit target labels, compared with 14/28 for first-token scoring. This result concerns candidate identification; it does not show general task understanding.

Shared-KV with an explicit \texttt{ABSTAIN} candidate remained unreliable when none of the task candidates was valid. On 60 held-out cases balanced between valid and no-valid-candidate inputs, the \texttt{ABSTAIN} option rejected 28/30 invalid cases but selected the correct non-abstaining value in only 17/30 valid cases. Autoregressive generation achieved 51/60 correct decisions, compared with 45/60 for Shared-KV with \texttt{ABSTAIN}. Candidate scores therefore cannot be treated as reliable confidence values, and safe rejection remains unresolved.

\subsection{Closed-Loop Manipulation and Verification}\label{closed-loop-manipulation-and-verification}

External simulator postconditions prevented model completion outputs from terminating closed-loop manipulation when the required simulator state was absent. SharedKV-BT completed 30/30 Stack, 21/30 full PickPlace, 26/30 NutAssemblySquare, and 25/30 Door episodes, for 102/120 overall. The execution rule blocked seven of eight completion requests because the relevant simulator postcondition was false; none ended the episode prematurely. These results verify that external simulator postconditions, rather than model completion outputs, held termination authority; they do not evaluate a separately trained verifier.

\begin{figure}[t]
\centering
\includegraphics[width=0.92\columnwidth]{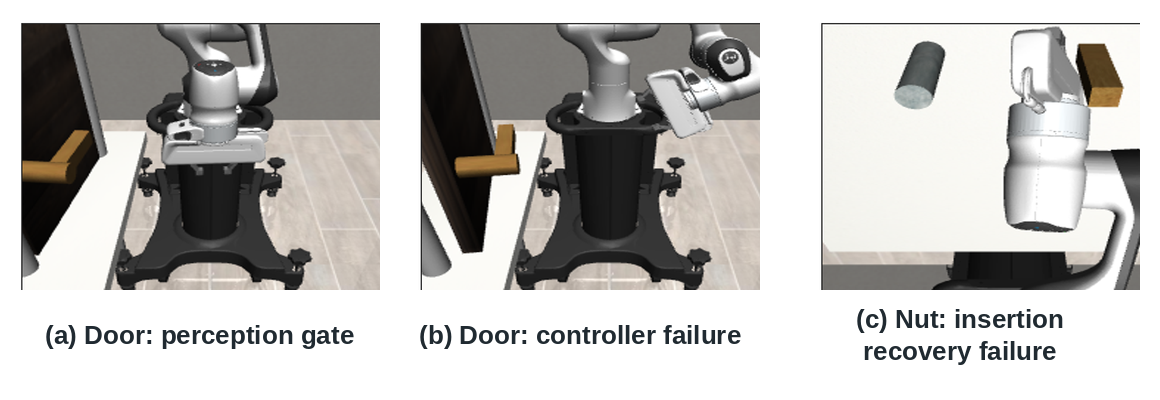}
\caption{Representative archived failure modes, not additional trials. (a) Door perception gate: incomplete evidence that the handle was visible caused the safety rule to withhold action. (b) Door controller failure: repeated permitted \texttt{OPEN\_DOOR} calls failed to open the door physically. (c) NutAssemblySquare insertion-recovery failure: recovery exhausted its attempt budget before insertion succeeded.}
\label{fig:failures}
\end{figure}

The 18 unsuccessful episodes identify downstream execution as the remaining limitation. Four NutAssemblySquare and five Door episodes reached their physical attempt limits, while nine full-PickPlace episodes ended after a pick-and-place skill failed. No broad-manipulation episode failed because of an invalid model output or an accepted premature completion. The archived cases in Fig.~\ref{fig:failures} distinguish conservative action withholding from repeated physical-skill failure.

\section{Discussion}\label{discussion}

The results distinguish three roles within SharedKV-BT. Shared-KV reduces the cost of typed decisions over predefined candidates, the active BT node focuses the decision interface, and external postconditions retain authority over progression. This separation keeps model selection, procedural control, execution, and verification attributable to different modules.

For \textbf{efficiency}, Shared-KV avoids recomputing a common task and state description for each predefined candidate. Its benefit grows when encoding the shared input accounts for most computation and shrinks when the candidate set is small or candidate labels are long (Fig.~\ref{fig:kvreuse}). Prompt-matched constrained AR confirms that the end-to-end speedup is not solely due to permitting invalid free-form outputs, although greedy tokenwise decoding selected different valid candidates on two inputs.

For \textbf{focus}, Adaptive PickPlace shows that the BT changes what the model must consider, not only when it decides (Fig.~\ref{fig:adaptive}). The node-local interface improved joint typed decisions and closed-loop execution when the global interface included valid but stage-irrelevant skills and arguments. Node-local accuracy remained higher under all seven tested candidate orders, although both interfaces remained order-sensitive on Acquire states. The evidence therefore supports BT nodes as boundaries for routing decisions to stage-specific interfaces, while also showing that likelihood-based candidate scoring remains order-sensitive. Masking stage-invalid choices in the global interface would provide the same stage-specific routing effect as the node-local condition.

For \textbf{authority}, the fixed-score policy replay showed that stage gating kept actions in order and external postconditions prevented premature completion. The closed-loop manipulation study likewise blocked seven completion requests for which the required simulator state was absent. Valid decisions still did not guarantee task completion when downstream navigation or manipulation failed. These results support using external postconditions to govern progression, while keeping decision and execution failures separately attributable.

\section{Limitations}\label{limitations}

The experiments cover simulated robot domains and three fixed computer-use tasks; they do not establish generalization to unseen tasks, physical hardware, or real-world perception. Recovery was tested with one injected planner failure. The Qwen3 check uses the same replay of saved pre-execution states and only ten closed-loop seeds.

Adaptive PickPlace uses manually specified structured states and a deterministic intervention schedule, does not test continuous geometric reasoning, and repeats states across seeds rather than providing 120 independent samples. The candidate-order sensitivity analysis is limited to seven fixed permutations and confirms that predictions vary with candidate order. Node-local remained better across these orders, but closed-loop robustness to order was not tested; the 18/30 versus 0/30 result uses only the published order.

SharedKV-BT also assumes predefined BTs, application-defined candidate sets, compatible field combinations, skills, and external postconditions. Reliable rejection when no candidate is valid and automatic compatibility enforcement remain unresolved; the \texttt{ABSTAIN} experiment indicates that candidate scores alone are insufficient for rejection. Some external postconditions and inputs use internal simulator state that would not be directly available to a real robot. The constrained AR baseline uses Python-level token-trie masking rather than an optimized grammar decoder. Finally, the measured speedups depend on the tested models, hardware, prompt lengths and structures, candidate-set sizes, and implementations.
\section{Conclusion}\label{conclusion}

SharedKV-BT separates three functions that agent systems often combine: Shared-KV makes typed decisions over predefined candidates faster, BT nodes focus the active interface, and external postconditions retain progression authority. Across the tested robot and computer-use protocols, this design reduced decision latency, prevented out-of-order or premature progression, and kept downstream execution failures distinct from model-selection errors.

\bibliographystyle{IEEEtran}
\bibliography{references}
\end{document}